\documentclass[11pt]{article}
\usepackage{eacl2017}
\usepackage{times}
\usepackage{url}
\usepackage{latexsym}
\usepackage{graphicx}
\usepackage{multirow}
\usepackage[english]{babel}
\usepackage[utf8x]{inputenc}
\usepackage{xcolor}

\eaclfinalcopy % Uncomment this line for the final submission
\title{Domain specialization: a post-training domain adaptation \\for Neural Machine Translation}

\author{
  Christophe Servan \ \  {\normalfont and} \ \  Josep Crego  \ \ {\normalfont and} \ \ Jean Senellart \\
  {\tt firstname.lastname@systrangroup.com} \\
  SYSTRAN / 5 rue Feydeau, 75002 Paris, France \\
}

\date{}

\begin{document}
\maketitle
\begin{abstract}
Domain adaptation is a key feature in Machine Translation. 
It generally encompasses terminology, domain and style adaptation, especially for human post-editing workflows in Computer Assisted Translation (CAT). 
With Neural Machine Translation (NMT), we introduce a new notion of domain adaptation that we call ``specialization'' and which is showing promising results both in the learning speed and in adaptation accuracy.
In this paper, we propose to explore this approach under several perspectives.
\end{abstract}

\section{Introduction}

Domain adaptation techniques have successfully been used in Statistical Machine Translation. 
It is well known that an optimized model on a specific genre (litterature, speech, IT, patent...) obtains higher accuracy results than a ``generic'' system. 
The adaptation process can be done before, during or after the training process. 
% Our preliminary experiments follow the latter approach. 

We propose to explore a new post-process approach, which incrementally adapt a ``generic'' model to a specific domain by running additional training epochs over newly available in-domain data.

In this way, adaptation proceeds incrementally when new in-domain data becomes available, generated by human translators in a post-edition context. 
Similar to the Computer Assisted Translation (CAT) framework described in \cite{Cettolo2014}.
\paragraph{Contributions}
The main contribution of this paper is a study of the new ``specialization'' approach, which aims to adapt generic NMT model without a full retraining process. 
% on the whole corpus.
Actually, it consist in using the generic model in a retraining phase, which only involves additional in-domain data.
Results show this approach can reach good performances in a far less time than full-retraining, which is a key feature to adapt rapidly models in a CAT framework.

% Our preliminary results show that incremental adaptation is effective for even limited amounts of in-domain data (less than 50k additional words). Constrained to use the original ``generic'' vocabulary, adaptation of the models can be run in a few seconds, showing clear quality improvements on in-domain test sets.

\section{Approach}

% Generally, when we seek to have a domain specific model, we train the model with specific data.
% % or , which belong to the desire domain.
% % But this process implies to train the model on the whole data.
% Within the framework of a Computer Assisted Translation (CAT) process, most of the time one cannot adapt or retrain a model on the whole generic and specific data, because of the time constraint.

Following the framework proposed by \cite{Cettolo2014}, we seek to adapt incrementally a generic model to a specific task or domain.
They show incremental adaptation brings new information in a Phrase-Based Statistical Machine Translation like terminology or style, which can also belong to the human translator.
Recent advances in Machine Translation focuses on Neural Machine Translation approaches, for which we propose a method to adapt incrementally to a specific domain, in this specific framework.

The main idea of the approach is to specialize a generic model already trained on generic data.
Hence, we propose to retrain the generic model on specific data, though several training iterations (see figure \ref{fig:selection}). 
The retraining process consist in re-estimating the conditional probability $p(y_1, \ldots , y_{m} |x_1, \ldots , x_n )$ where $(x_1, \ldots , x_n )$ 
is an input sequence of length $n$ and $(y_1, \ldots , y_{m})$ is its corresponding output sequence whose length $m$ may differ from $n$. This is done without dropping the previous learning states of the Recurrent Neural Network. 

The resulting model is considered as adapted or specialized to a specific domain.
% Figure \ref{fig:selection} presents our approach.

% In addition to this specialization process, we propose to used the bivec model, a bilingual representation \cite{Luong2015effective}, trained on the in-domain parallel corpus to handle Out-of-Vocabulary words while translating.

% The main advantage of such approach is to reduce the 

% by picking out a subset of training data that are most relevant 
% to the domain of interest.

\begin{figure}[t!]
\begin{center}
%\hspace{-.7cm}
\includegraphics[width=\linewidth]{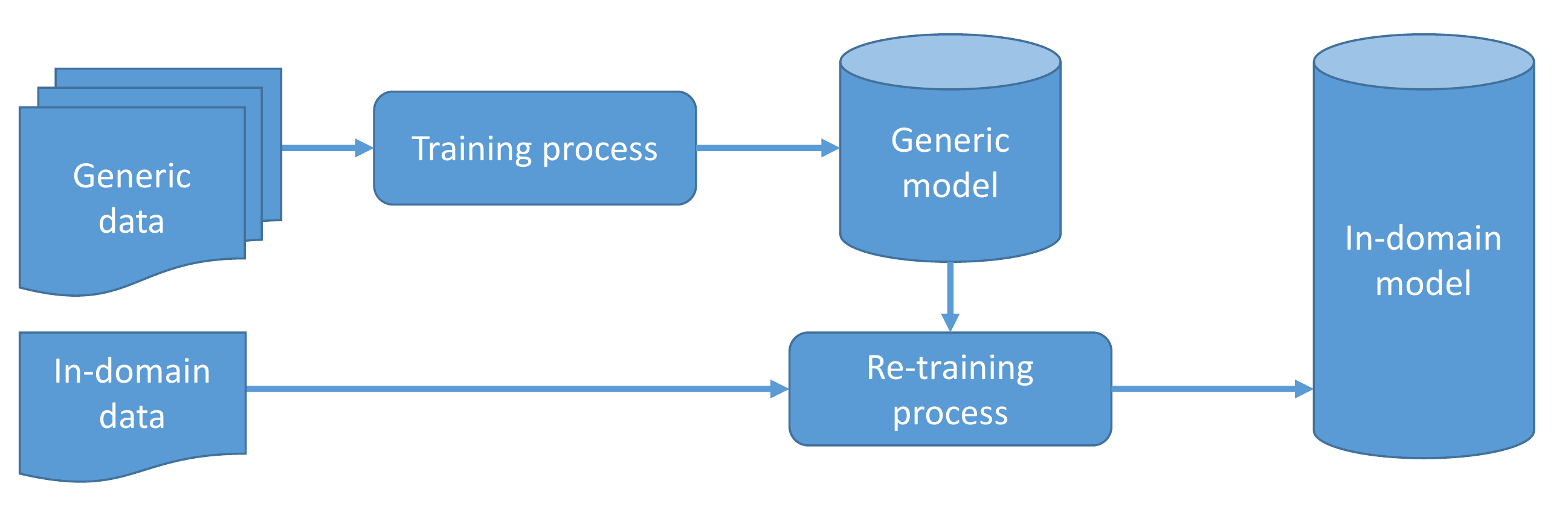}
\end{center}
\caption{\label{fig:model}The generic model is trained with generic data, then the generic model obtained is retrained with in-domain data to generate an specialized model.}
\label{fig:selection}
\end{figure}

\section{Experiment framework}

% This section presents our first expriments done using this approach.

We create our own data framework described in the next section and we evaluate our results using the BLEU score \cite{papineni02bleu} and the TER \cite{Snover2006}.

The Neural Machine Translation system combines the attention model approach \cite{Luong2015effective} jointly with the sequence-to-sequence approach \cite{Sutskever2014}.
% It also includes a case feature, which is collated in the input layer to the embedding see figure \ref{fig:feature_embedding}.
% 
% \begin{figure}
%  \includegraphics[width=\linewidth]{Linguisticfeature}
%  \caption{\label{fig:feature_embedding} The input layer, composed of the word embeddings and the feature embeddings.}
% \end{figure}

According to our approach, we propose to compare several configurations, in which main difference is the training corpus. 
On one hand, we consider the generic data and several amounts of in-domain data for the training process.
On the other hand, only the generic data are considered for the training process, then several amounts of in-domain data are used only for the specialization process in a retraining phase.
The main idea behind these experiment is to simulate an incremental adaptation framework, which enables the adaptation process only when data are available (e.g.: translation post-editions done by a human translator.)

The approach is studied in the light of two experiments and a short linguistic study.
The first experiment concerns the impact of ``specialization'' approach among several additional epochs; then, the second one, focuses on the amount of data needed to observe a significant impact on the translation scores.
Finally, we propose to compare some translation examples from several outputs.

\subsection{Training data}
\label{sec:data}
The table \ref{tab:alldata} presents all data used in our experiments. 
We propose to create a generic model with comparable amount of several corpora, which each of them belong to a specific domain (IT, literature, news, parliament). 
All corpora are available from the OPUS repository \cite{Tiedemann2012parallel}.

We propose to specialize the generic model using a last corpus, which is a corpus extracted from the European Medical Agency (\textit{emea}).
The corpus is composed of more than 650 documents, which are medicine manuals. 

We took apart a $2K$ lines as test corpus, then, to simulate the incremental adding of data, we created four training corpora corresponding to several amount of documents: $500$, $5K$, $50K$ and all the lines of the training corpus.
These amount of data are corresponding roughly to $10\%$ of a document, one document and ten documents, respectively.
% We did the same with an in-house dialogue (\textit{dial}) corpus, which we combined with the TED corpus.

% Data used in our experiments have been preprocessed with the \textit{byte pair encoding} compression algorithm \cite{sennrich2016improving} and are presented in next section.

% All the corpora were pre-processed with the \textit{byte pair encoding} compression algorithm \cite{sennrich2016improving} with 30K operations, to avoid Out-of-Vocabulary and new in-domain vocabulary words.
\begin{table}
\begin{center}
\small
\begin{tabular}{llrrr}
\hline
Type & Domain & \#lines & \#src tokens & \#tgt tokens \\
\hline
\multirow{4}{*}{Train}& \textit{generic} & 3.4M & 73M & 86M \\
\cline{2-5}% \hline
& \textit{emea-0.5K} & 500 & 5.6K & 6.6K \\
& \textit{emea-5K} & 5K & 56.1K & 66.4K \\
% & \textit{emea-25K} & 25K & 280K & 331K \\
& \textit{emea-50K} & 50K & 568K & 670K \\
% & \textit{emea-250K} & 250K & 2.8M & 3.3M \\
% & \textit{emea-500K} & 500K & 5.6M & 6.7M \\
& \textit{emea-full} & 922K & 10.5M & 12.3M \\
% \cline{2-5}
% & \textit{dial}-25k & 25K & 225K & 221K \\
% & \textit{dial}-250k & 250K & 2.2M & 2.2M \\
% & \textit{dial}-500k & 500K & 4.5M & 4.4M \\
% & \textit{dial}-full & 1.7M & 152M & 150M \\
\hline
% \multirow{3}{*}{Test} 
dev.& \textit{generic} & 2K& 43.7K & 51.3K \\
\hline
test & \textit{emea} & 2K & 35.6K & 42.9K  \\
% \cline{2-5}
% & \textit{dial} & 2K & 24.6K & 25K \\
\hline
\end{tabular}
\end{center}
\caption{\label{tab:alldata} details of corpora used in this paper.}
\end{table}

% emea_500.en.gz    500    5634
% emea_500.fr.gz    500    6682
% emea_5000.en.gz   5000   56131
% emea_5000.fr.gz   5000   66468
% emea_50000.en.gz  50000  568408
% emea_50000.fr.gz  50000  670511

\subsection{Training Details}
\label{ssec:training}

The Neural Machine Translation approach we use is following the sequence-to-sequence approach \cite{Sutskever2014} combined with attentional architecture \cite{Luong2015effective}.
In addition, all the generic and in-domain data are pre-processed using the \textit{byte pair encoding} compression algorithm \cite{sennrich2016improving} with 30K operations, to avoid Out-of-Vocabulary words.

We keep the most frequent $32K$ words for both source and target languages with $4$ hidden layers with $500$-dimensional embeddings and $800$ bidirectional Long-Short Term Memory (bi-LSTM) cells.
% size of $500$ cells and $4$ hidden layers which are composed of $800$ cells each. 
During training we use a mini-batch size of $64$ with dropout probability set to $0.3$.
% stochastic gradient descent, 
We train our models for $18$ epochs and the learning rate is set to $1$ and start decay after epoch $10$ by $0.5$.
It takes about $8$ days to train the generic model on our NVidia GeForce GTX 1080.
% using the \textit{generic} data.

The models were trained with the open-source toolkit \texttt{seq2seq-attn}\footnote{\url{https://github.com/harvardnlp/seq2seq-attn}} \cite{kim2016sequence}.

\subsection{Experiments}
\begin{table}
\small
\centering
 \begin{tabular}{p{2.7cm}cc}
 \hline
 Models & BLEU & TER\\
 \hline
\textit{generic}& 26.23 & 62.47 \\
\textit{generic}$+$\textit{emea-0.5K} & 26.48 & 63.09\\
\textit{generic}$+$\textit{emea-5K} & 28.99 & 58.98\\
\textit{generic}$+$\textit{emea-50K} & 33.76& 53.87\\
\textit{generic}$+$\textit{emea-full} & 41.97 & 47.07\\
 \hline
 \end{tabular}
\caption{\label{tab:resultsFull} BLEU score of full trained systems.}
\end{table}

As a baseline, we fully trained five systems, one with the generic data (\textit{generic}) and the other with generic and various amount of in-domain data: $500$ lines (\textit{emea-0.5K}), $5K$ lines (\textit{emea-5K}) and $50K$ lines (\textit{emea-50K}).
The evaluation is done on the in-domain test (\textit{emea-tst}) and presented in the table \ref{tab:resultsFull}. 
Without surprises, the more the model is trained with in-domain data the more BLEU and TER scores are improved.
These models are baselines in incremental adaptation experiments.

\subsubsection{Performances among training iterations}
\label{sec:stuEpochs}
\begin{figure}
\begin{center}
%\hspace{-.7cm}
\includegraphics[width=\linewidth]{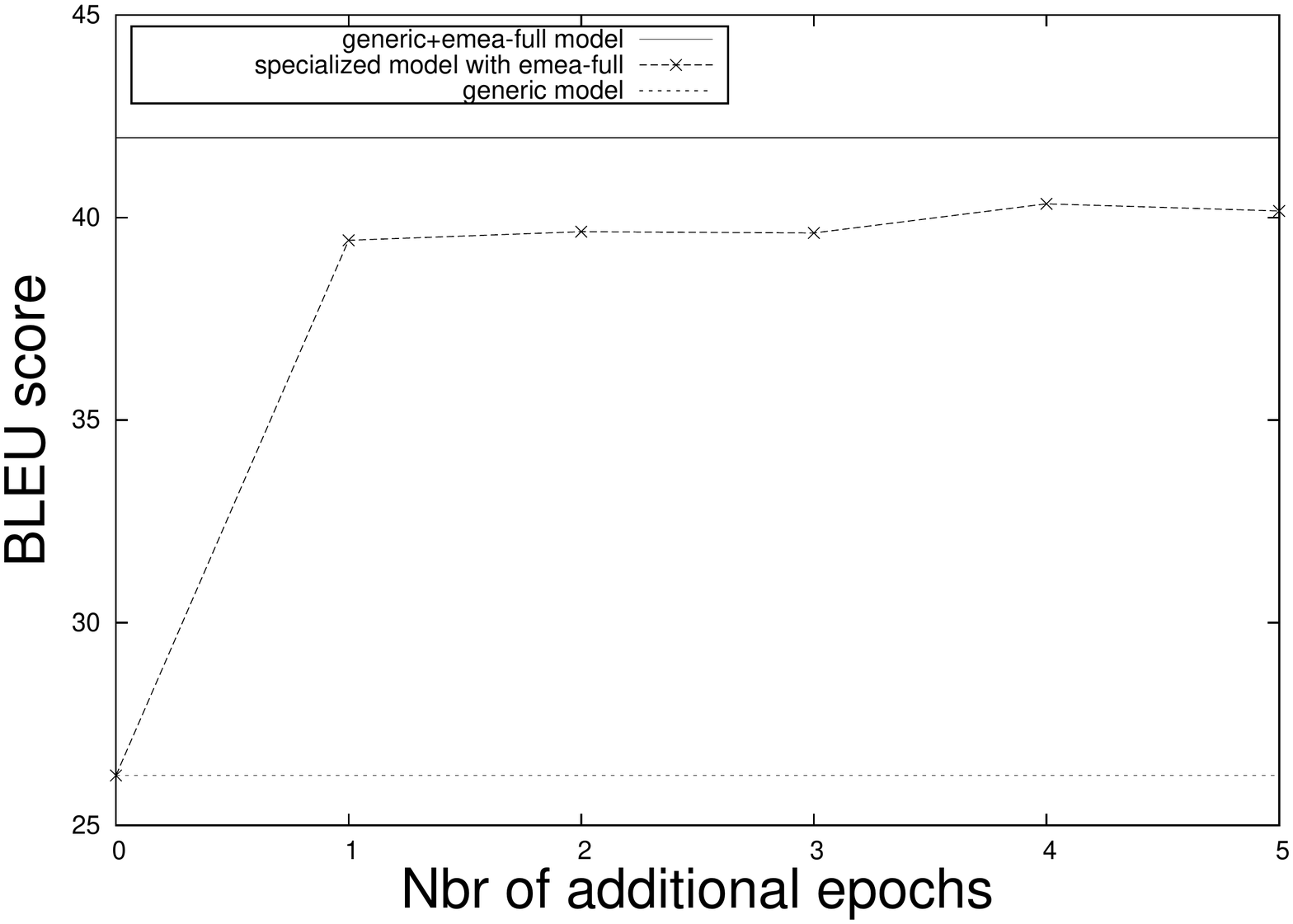}
\end{center}
\caption{\label{fig:incEpohcs}Curve of ``specialization'' performances among epochs.}
\label{fig:selection}
\end{figure}

The first study aims to evaluate the approach among additional training iterations (also called ``epochs''). 
Figure \ref{fig:incEpohcs} presents the curve of performances when the specialization approach is applied to the \textit{generic} model by using all in-domain data (\textit{emea-full}). 

We compare the results with two baselines: on the top of the graphic, we show a line which corresponds to the score obtained by the model trained with both generic and in-domain data (noted \textit{generic}$+$\textit{emea-full}); 
On the bottom, the line is associated to the generic model score, which is trained with only generic data (noted \textit{generic}).
The curve is done with the generic model specialized with five epochs additional epochs on all in-domain data (noted \textit{specialized model with emea}).
In the graphic, we can observe that a gap obtained with the first additional epoch with more than $13$ points, but then the BLEU score improves around $0.15$ points with each additional epoch and tend to stall after $10$ epochs (not shown).

So far, the specialization approach does not replace a full retraining, while the specialization curve does not reach the \textit{generic}$+$\textit{emea-full} model. 
But, the retraining time of one additional epoch with all in-domain data is around $1$ hour and $45$ minutes, while a full retraining would takes more than $8$ days.

In our CAT framework, even $1$ hour and $45$ minutes is too much, the adaptation process need to be performed faster with smaller amount of data like a part of a document ($500$ lines) or a full document ($5K$ lines). 
Considering the time constraint, the approach tends to be performed though one additional epoch. 
% in this framework
\subsubsection{Performances among data size}

The second experiment concerns the observation of specialization performances when we vary the amount of data.
Using the data presented Table \ref{tab:alldata}, we apply the specialization process on the generic corpus by taking $0.5K$, $5K$, $50K$ and all the in-domain data (as presented in section \ref{sec:data}).
According to our previous study (see section \ref{sec:stuEpochs}), we focuses on the results obtained with only one additional epoch.

\begin{table}
\small
\centering
 \begin{tabular}{p{2.7cm}p{1.8cm}cc}
 \hline
 Training corpus & Specialization corpus & BLEU & TER\\
 \hline
\textit{generic}& N/A & 26.23 & 62.47 \\
\textit{generic}$+$\textit{emea-0.5K} & N/A &  26.48 & 63.09\\
\textit{generic}$+$\textit{emea-5K} & N/A &  28.99 & 58.98\\
\textit{generic}$+$\textit{emea-50K} & N/A &   33.76& 53.87\\
\textit{generic}$+$\textit{emea-full} & N/A & 41.97 & 47.07 \\
 \hline
\textit{generic}& \textit{emea-0.5K} & 27.33 & 60.92 \\
\textit{generic}& \textit{emea-5K} & 28.41 & 58.84\\
% \textit{generic}& \textit{emea}-25K & 31.33 & 55.99\\
\textit{generic}& \textit{emea-50K} & 34.25 & 53.47\\
% \textit{generic}& \textit{emea}-250K & 37.57 & 50.67\\
% \textit{generic}& \textit{emea}-500K & 38.82 & 50.23\\
\textit{generic}& \textit{emea-full} & 39.44 & 49.24 \\
 \hline
 \end{tabular}
\caption{\label{tab:results} BLEU and TER scores of the specialization approach on the in-domain test set.}
% , according to the . This table presents on the top the performances of full retrained models, which can be compared with the specialized models with the same amount of data.}
\end{table}

\begin{table}
\scriptsize
\centering
 \begin{tabular}{llrrrrr}
\hline
Process & Corpus & \#lines & \#src & \#tgt & Process \\
& & & tokens & tokens & time\\
\hline
Train& \textit{generic} & 3.4M & 73M & 86M & 8 days\\
\hline
& \textit{emea-0.5K} & 500 & 5.6K & 6.6K & $<$1 min\\
Speciali-& \textit{emea-5K} & 5K & 56.1K & 66.4K & $\approx$1 min\\
zation& \textit{emea-50K} & 50K & 568K & 670K & $\approx$6 min\\
% 
% & \textit{emea-25K} & 25K & 280K & 331K & 4 min\\
% & \textit{emea-250K} & 250K & 2.8M & 3.3M & 29 min\\
% & \textit{emea-500K} & 500K & 5.6M & 6.7M & 59 min\\
& \textit{emea-full} & 922K & 10.5M & 12.3M & 105 min \\
\hline
 \end{tabular}
\caption{\label{tab:timeresults} Time spent for each process, the training and the specialization process, according to the amount of data we have.}
\end{table}

\begin{table*}[h!]
\centering
\small
 \begin{tabular}{ll}
\hline
Source: & What benefit has SonoVue shown during the studies ? \\
\hline
Reference: & Quel est le bénéfice démontré par SonoVue au cours des études ? \\
\hline
\textit{generic model} & Quel {\color{red} avantage SSonVue} {\color{blue!50!black}a-t-il montré pendant les études} ? \\
\hline
specialization \textit{emea-0.5K} & Quel {\color{green!50!black} bénéfice} {\color{red} SSonVue} {\color{blue!50!black}a-t-il montré lors des études} ? \\
specialization \textit{emea-5K} & Quel {\color{green!50!black} bénéfice} {\color{red} SSonVue a-il} {\color{blue!50!black}montré pendant les études} ? \\
specialization \textit{emea-50K} & {\color{red} Quels} {\color{green!50!black} est le bénéfice démontré par SonoVue au cours des études} ? \\
\hline
 \end{tabular}
\caption{\label{tab:ex} Example of translation output of the generic model and the specialized models with different amount of in-domain data. Red, blue and green are, respectively, \textit{\color{red}bad}, \textit{\color{blue!50!black}acceptable} and \textit{\color{green!50!black}good} translations.}
\end{table*}

We can observe that with only 500 lines, the improvements reaches more than $1$ BLEU points and $2$ TER points. 
Then, with 10 time more additional data, BLEU and TER scores improved the baseline of $2$ and nearly $4$ points, respectively.
With more additional data ($10$ documents), improvements reach $8$ points of BLEU and $9$ points of TER.
Finally with all the in-domain data available, the specialization increase the baseline of $13$ points of both BLEU and TER scores.

Comparing the approach with retraining all the generic data, with the same amount of in-domain data, it appears our approach reaches nearly the same results. 
Moreover, with $50K$ of in-domain data, the specialization approach performs better of $0.5$ of BLEU and TER points.
But, when we have much more in-domain data available, the specialization approach does not outperforms the full retraining ($39.44$ against $41.97$ BLEU points).

\subsection{Discussion}

% It seems this approach is also stalling accord to the amount of data added. But the time spend to train an additionnal epoch vary according to the amount of data we add.
Focussing on the time constraint of the CAT framework, the table \ref{tab:timeresults} presents the time taken to process our specialization approach. It goes from less than one minutes to more than $1$ hour and $45$ minutes.
If we compare this table with the table \ref{tab:results}, we observe that this approach enables to gain $1$ BLEU point in less than $1$ minute, $2$ points in $1$ minute and more than $6$ BLEU points in $6$ minutes.
The ratio of "time spent" to "score gained" seems impressive.
% , especially if we want to quickly adapt models within a CAT framework.

The table \ref{tab:ex} shows an example of the outputs obtained with the specialization approach.
We compare the generic model compared to the specialized models with respectively $0.5K$, $5K$ and $50K$ lines of in-domain data. 

We can clearly see the improvements obtained on the translation outputs. 
Even if the last one does not stick strictly to the reference, the translation output can be considered as a good translation (syntactically well formed and semantically equivalent).

This specialization approach can be seen as an optimization process (like in classical Phrase-Based approach), which aims to tune the model \cite{Och2003}.

\section{Related work}

Last years, domain adaptation for machine translation has received lot of attention and studies.
These approaches can be processed at three levels: the pre-processing, the training, the post-processing.
In a CAT framework, most of the approaches focuses on the pre-processing or on the post-processing to adapt models.

Such pre-processing approaches like data selection introduced by \cite{Lue2007} and improved by \cite{Gao2002improving} and many others \cite{moore2010dataSelection,Axelrod2011} are effective and their impact studied \cite{Lambert2011,Cettolo2014,Wuebker2014}. 
But, the main draw back of these approaches is they need a full retrain to be effective.

% During the training, several approaches propose to give some domain features to the model. 
% But theses approach need a full training process.

The post-training family concerns methods which aims to update the model or to optimize the model to a specific domain. 
Our approach belongs to this category.

This approach is inspired by \cite{Luong2015}, they propose to train a generic model and, then, they further a training over a dozen of epochs on a full in-domain data (the TED corpus).
We do believe this approach is under estimated and we propose to study its efficiency in a specific CAT framework with a few data. 
On one hand, we propose to follow this approach by proposing to use a fully trained generic model. 
But, on the other hand, we propose to train further only on small specific data over a few additional epochs (from 1 to 5).
In this way, our approach is slightly different and can be equated to a tuning process \cite{Och2003}.
% This paper is focused on the study of this approach by varying the amount of in-domain data used to retrain the generic model

\section{Conclusion}

In this paper we propose a study of the ``specialization'' approach.
This domain adaptation approach shows good improvements with few in-domain data in a very short time. 
For instance, to gain 2 BLEU points, we used 5K lines of in-domain data, which takes 1 minute to be performed.

Moreover, this approach reaches the same results as a full retraining, when $10$ documents are available. 
Within a CAT framework, this approach could be a solution for incremental adaptation of NMT models, and could be performed between two rounds of post-edition.
In this way, we propose as future work to evaluate our approach in a real CAT framework.
% , by adapting from the model to the document the translator is post-editing.

\bibliography{eacl2017}
\bibliographystyle{eacl2017}

% \appendix
% 
% \section{Supplemental Material}
% \label{sec:supplemental}
% EACL-2017 also encourages the submission of supplementary material
% to report preprocessing decisions, model parameters, and other details
% necessary for the replication of the experiments reported in the 
% paper. Seemingly small preprocessing decisions can sometimes make
% a large difference in performance, so it is crucial to record such
% decisions to precisely characterise state-of-the-art methods.
% 
% Nonetheless, supplementary material should be supplementary (rather
% than central) to the paper. It may include explanations or details
% of proofs or derivations that do not fit into the paper, lists of
% features or feature templates, sample inputs and outputs for a system,
% pseudo-code or source code, and data. (Source code and data should
% be separate uploads, rather than part of the paper).
% 
% The paper should not rely on the supplementary material: while the paper
% may refer to and cite the supplementary material will be available to the
% reviewers, they will not be asked to review the
% supplementary material.
% 
% Appendices (i.e. supplementary material in the form of proofs, tables,
% or pseudo-code) should come after the references, as shown here. Use
% \verb|\appendix| before any appendix section to switch the section
% numbering over to letters.
% 
% \section{Multiple Appendices}
% \dots can be obtained by using more than one section. We hope you won't
% need that.
% 

\end{document}